\UseRawInputEncoding
\documentclass[a4paper,11pt]{article}

\usepackage{fourier}
\usepackage{color}
\usepackage{graphicx}
\usepackage{url}
\usepackage[affil-it]{authblk}
\usepackage{amsmath}
\usepackage{wrapfig}

\usepackage[T1]{fontenc}
\usepackage{times}

\usepackage{hyperref}
\begin{document}

\title{Learning from Exemplary Explanations}

\author[1,2]{Misgina Tsighe Hagos}
\author[1,3]{Kathleen M. Curran}
\author[1,2]{Brian Mac Namee}
\affil[1]{Science Foundation Ireland Centre for Research Training in Machine Learning}
\affil[2]{School of Computer Science, University College Dublin}
\affil[3]{School of Medicine, University College Dublin}

\date{}
\maketitle
\thispagestyle{empty}

\begin{abstract}
eXplanation Based Learning (XBL) is a form of Interactive Machine Learning (IML) that provides a model refining approach via user feedback collected on model explanations. Although the interactivity of XBL promotes model transparency, XBL requires a huge amount of user interaction and can become expensive as feedback is in the form of detailed annotation rather than simple category labelling which is more common in IML. This expense is exacerbated in high stakes domains such as medical image classification. To reduce the effort and expense of XBL we introduce a new approach that uses two input instances and their corresponding Gradient Weighted Class Activation Mapping (GradCAM) model explanations as exemplary explanations to implement XBL. Using a medical image classification task, we demonstrate that, using minimal human input, our approach produces improved explanations ($+0.02, +3\%$) and achieves reduced classification performance ($-0.04, -4\%$) when compared against a model trained without interactions.
\end{abstract}
\textbf{Keywords:} Explanation based Learning, Interactive Learning, Medical Image Classification. 


\section{Introduction}

\begin{figure}[h]
\vskip 0.2in
\begin{center}
\centerline{\includegraphics[width=0.65\linewidth]{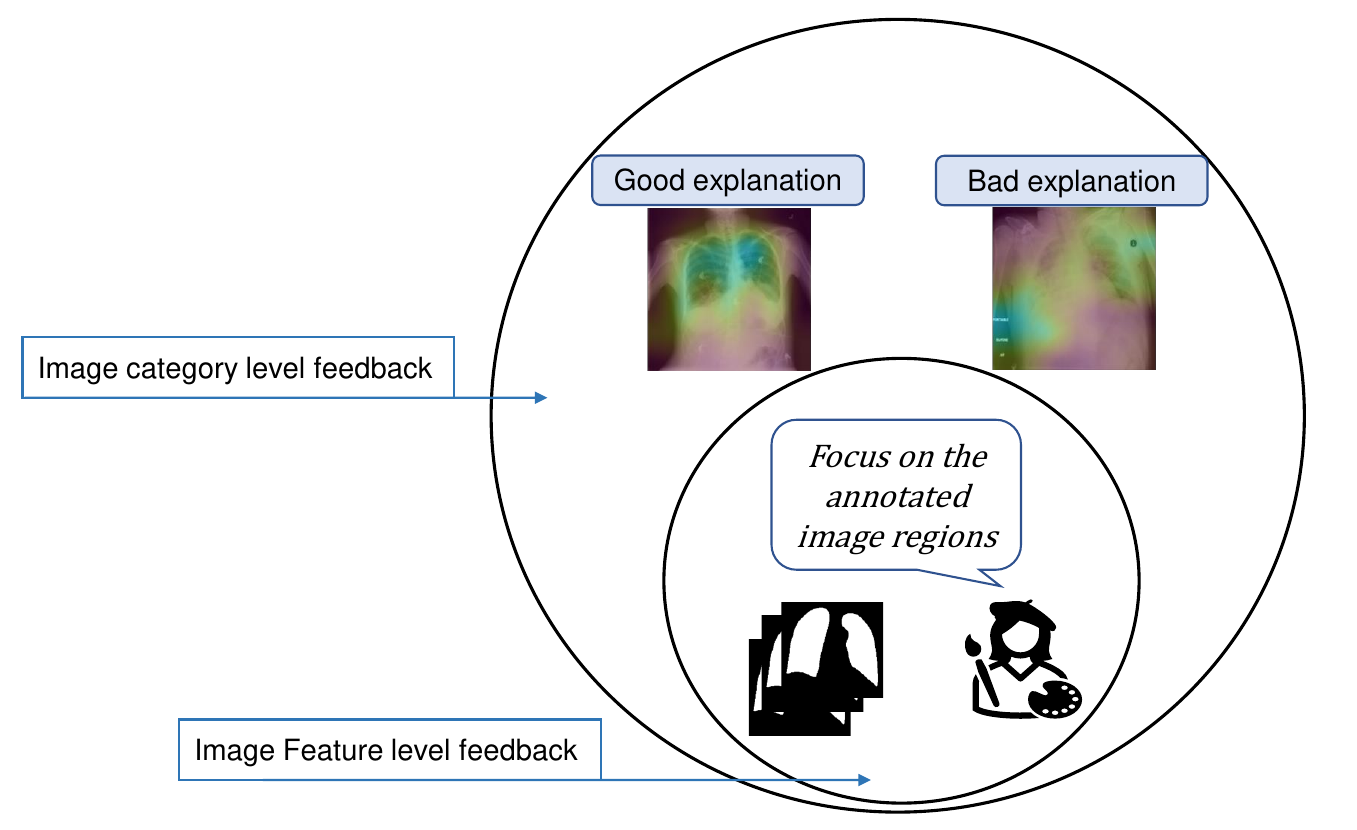}}
\vspace{-10pt}
\caption{The inner circle shows the typical mode of feedback collection where users annotate image features. The outer circle shows how the Exemplary eXplanation Based Learning (eXBL) approach requires only identification of one good and one bad explanation.}
\label{figure:feedback_abstraction_in_xbl}
\end{center}
\vspace{-20pt}
\end{figure}

Interactive Machine Learning (IML) is an approach that aims to provide a platform for user involvement in the model training or retraining process \cite{fails2003interactive}. The literature on IML is dominated by active learning which reduces the manual effort associated with creating labelled training datasets by interactively selecting a sub-sample of an unlabelled dataset for manual labelling \cite{budd2021survey}. However, eXplanation Based Learning (XBL) has recently begun to gain traction as it allows deeper interaction with users by providing an opportunity to collect feedback on model explanations \cite{stumpf2009interacting,kulesza2015principles,10.3389/frai.2023.1066049}. This form of interaction allows a more transparent form of model training than other IML approaches as users get a chance to refine a model by interacting-with and correcting its explanations.

XBL starts off with a learner model, $f$, that was initially trained using a simple classification loss, categorical cross entropy for example, which is calculated based on the error between the model's prediction and ground-truth label. Then, XBL typically refines $f$ by augmenting its classification loss with an explanation loss,

\begin{align}
\label{formula:generic_xbl}
    L = L_{CE} + L_{expl} + \lambda{}\sum_{i=0}\theta_i
\end{align}

In Equation (\ref{formula:generic_xbl}), $L_{CE}$ is the traditional categorical cross entropy which is calculated based on the error between the model's predictions and ground-truth labels; $L_{expl}$ is an explanation loss that is computed between the explanation produced from a model and a manual annotation of input instances, $M$; $\lambda$ is a regularisation term used to avoid overfitting that could be caused by the introduction of the new loss term, $L_{expl}$; and $\theta$ refers to network parameters. $M$ can be a mask showing the important image regions that a learner should focus on or a mask of confounding or non-salient regions that a model should ignore. Saliency based feature attributions are usually used to generate model explanations. One example, from \cite{schramowski2020making} formulates the explanation loss for training instances $x \in X$ of size N and Gradient Weighted Class Activation Mapping (GradCAM) model explanations generated using a trained model $f$ as shown in Equation (\ref{equation:schramowski_gradcam}). GradCAM is a saliency based local model explanation technique \cite{selvaraju2017grad}.

\begin{align}
\label{equation:schramowski_gradcam}
    L_{expl} = \sum_{i=0}^N M_{i}GradCAM(x_{i})
\end{align}

As is seen in the inner circle of Figure \ref{figure:feedback_abstraction_in_xbl}, in XBL, the most common mode of user interaction is image feature annotation. This requires user engagement that is considerably much more demanding than the simple instance labelling that most IML techniques require \cite{zlateski2018importance} and increases the time and cost of feedback collection in XBL. As can be seen in the outer circle of Figure \ref{figure:feedback_abstraction_in_xbl}, we are interested in lifting this pressure from users (or feedback providers) and simplifying the interaction to ask for identification of two explanations as exemplary explanations and ranking them as good and bad explanations, and so make feedback collection cheaper and faster. This kind of user interaction where users are asked for a ranking instead of category labels has also been found to increase inter-rater reliability and data collection efficiency \cite{o2017rating}. We incorporate this feedback into model training through a contrastive loss; specifically, triplet loss \cite{chechik2010large}. 

The main goal of this paper is to demonstrate the effectiveness this loss based on just two exemplars. Therefore, we use an existing feature annotated dataset to identify good and bad explanations to demonstrate suitability of our proposal. In a real-world interactive learning scenario where end users have to choose the good and bad explanations, active learning approaches can be used to reduce the pool of explanations users have to choose the explanations from.

The main contributions of this paper are:

\begin{enumerate}
    \item We propose the first type of eXplanation Based Learning (XBL) that can learn from only two exemplary explanations of two training images;
    \item We adopt triplet loss for XBL to incorporate the two exemplary explanations into an explanation loss;
    \item In addition to showing that XBL can be implemented with just two instances, our experiments demonstrate that our proposed method achieves improved explanations and comparable classification performance when compared against a baseline model.
\end{enumerate}

\section{Related Work}
\label{section:related_work}

Based on the approach utilised to incorporate user feedback into model training, XBL methods can be generally categorised into two: (1) augmenting loss functions; and (2) augmenting training datasets using user feedback by removing confounding or spurious regions identified by users.

\paragraph{Augmenting Loss Functions.}

XBL methods that fall under this category follow the approach introduced in Equation \ref{formula:generic_xbl} by adding an explanation loss to a model's training to refine it to focus on image regions that are considered relevant by user(s) or to ignore confounding regions. One example of this category is Right for the Right Reasons (RRR) \cite{ross2017right} that penalises a model with high input gradient model explanations on the wrong image regions based on user annotation. It uses,

\begin{equation}
\label{equation:rrr}
    L_{expl} = \sum_{n}^N \left[ M_n \frac{\partial} {\partial x_n}\sum_{k=1}^K \log{\hat{y}_{nk}} \right] ^2
\end{equation}    
for a function $f(X|\theta)=\hat{y} \in \mathbb{R}^{N\times K} $ trained on images \begin{math}x_{n}\end{math} of size $N$ with $K$ categories, where \begin{math}M_{n} \in \ \{0,\ 1\}\end{math} is user annotation of image regions that should be avoided by the model.



Similarly, Right for Better Reasons (RBR) \cite{shao2021right} uses Influence Functions (IF) in place of input gradients to correct a model's behaviour. Contextual Decomposition Explanation Penalisation (CDEP) \cite{rieger2020interpretations} penalises features and feature interactions.

User feedback in XBL experiments can be either: (1) telling the model to ignore non-salient image regions; or (2) instructing the model to focus on important image regions in a training dataset \cite{hagos2022impact}. While the XBL methods presented above refine a model by using the first feedback type, Human Importance-aware Network Tuning (HINT) does the opposite by teaching a model to focus on important image parts using GradCAM model explanations \cite{selvaraju2019taking}.


\paragraph{Augmenting Training Dataset.}

In addition to augmenting loss functions, XBL can also be implemented by augmenting a  training dataset based on user feedback. Instance relabelling \cite{teso2021interactive}, counterexamples generation \cite{teso2019explanatory}, and using user feedback as new training instances \cite{popordanoska2020machine} are some of the methods that augment a dataset to incorporate user feedback into XBL.


While XBL approaches show promise in unlearning spurious correlations that a model might have learned by giving attention to non-relevant or confounding image regions \cite{hagos2022identifying,pfeuffer2023explanatory}, they all need a lot of effort from users. In order to unlearn spurious correlations from a classifier, \cite{pfeuffer2023explanatory} collected feature annotation on 3000 chest x-ray images. This kind of demanding task hinders practical deployment and domain transferability of XBL. For this reason, it is of paramount importance to build an XBL method that can refine a trained model using a limited amount of user interaction in order to achieve a plausible and domain transferable implementation. To the best of our knowledge, this area of XBL is completely unexplored.

\section{Exemplary eXplanation Based Learning}
\label{section:eXBL}

As is illustrated by Equations \ref{equation:schramowski_gradcam} and \ref{equation:rrr}, for typical XBL approaches, user annotation of image features, or $M$, is an important prerequisite. We introduce Exemplary eXplanation Based Learning (eXBL) to mitigate the time and resource complexity caused by the feature annotation process. In eXBL, we propose to simplify the expensive feature annotation requirement and replace it with two exemplary explanations: \emph{Good GradCAM explanation} ($C_{good}$) and \emph{Bad GradCAM explanation} ($C_{bad}$). However, even if this replaces feature annotation with two labels, categorising explanations would still be expensive if it's to be performed for all training instances whose size could be in the thousands. For this reason, we only use one $C_{good}$ and one $C_{bad}$. 

We choose to use GradCAM model explanations because they have been found to be more sensitive to training label reshuffling and model parameter randomisation than other saliency based explanations \cite{adebayo2018sanity}. To select the good and bad explanations from a list of generated GradCAM explanations, we use an objective explanation metric, called Activation Recall (AR). AR measures how much of the actual relevant parts of test images, $M$, are considered relevant by a model. While a larger AR value means a model is giving higher attention to relevant image regions, a smaller AR would mean the model is not focusing on relevant image parts for its prediction. AR is formulated as follows,

\begin{equation}
    AR_{x \in X} =  \frac{ GradCAM(x) * M}{M} 
\end{equation}

We then assign products of input instances and GradCAM explanation to $C_{bad}$ and $C_{good}$ using the instances with maximum and minimum AR values, as follows, 

\begin{equation}
    C_{good} := i\cdot GradCAM(i) , \max_{x\in X}(AR(x)) := AR_{i}
\end{equation}

\begin{equation}
    C_{bad} := j\cdot GradCAM(j) , \min_{x\in X}(AR(x)) := AR_{j}
\end{equation}

The product of the input instance and the Grad-CAM explanation is used instead of just the Grad-CAM explanation because taking only the GradCAM outputs to be the good/ bad explanations could lead to biased exemplary explanations as it would mean we are only taking the model's focus or attention into consideration.



We then take inspiration from triplet loss to incorporate $C_{good}$ and $C_{bad}$ into our explanation loss. The main purpose of our explanation loss is to penalise a trainer according to its distance from $C_{good}$ and $C_{bad}$: The closest to $C_{good}$ and the furthest from $C_{bad}$, the lower the loss.

For the product of the training instances $x\in X$, and their corresponding GradCAM outputs, $x \cdot GradCAM(x)$, we compute the euclidean distances $d_{xg}$ and $d_{xb}$, which represent distances from $C_{good}$ and $C_{bad}$ as follows,

\begin{equation}
    d_{xg} := d(x \cdot GradCAM(x), C_{good})
\end{equation}

\begin{equation}    
    d_{xb} := d(x \cdot GradCAM(x), C_{bad})
\end{equation}

We train the model $f$ to achieve $d_{xg} \ll d_{xb}$ for all $x$. We do this by adding a $margin = 1.0$; $d_{xg} - d_{xb} + margin < 0$.


We then compute the explanation loss as follows,

\begin{equation}
    \label{equation:proposed_lexp}
    L_{expl} = \sum_{i}^N \max(d_{x_{i}g} - d_{x_{i}b} + margin, 0) 
\end{equation}

In addition to correctly classifying the training images, which is achieved through $L_{CE}$, this $L_{expl}$ (Equation \ref{equation:proposed_lexp}) would train $f$ to output GradCAM values that resemble the good explanations and that differ from the bad explanations.

\begin{figure*}[t]
\vskip 0.2in
\begin{center}
\centerline{\includegraphics[width=1.\linewidth]{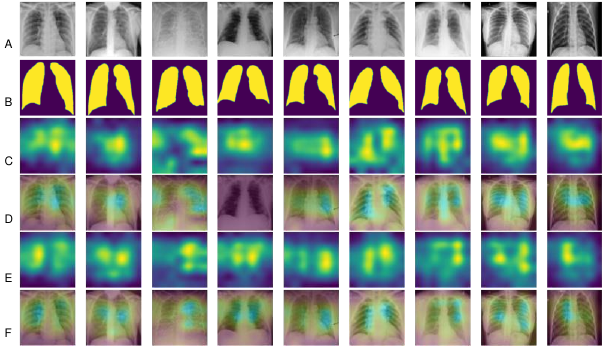}}
\caption{(A) Input images. (B) Feature annotations masks. (C) GradCAM explanations of the Unrefined model. (D) GradCAM outputs of Unrefined model overlaid over input images. (E) GradCAM explanations of eXBL. (F) GradCAM outputs of eXBL model overlaid over input images.}
\label{figure:sample_gradcam_results}
\end{center}
\vskip -0.2in
\end{figure*}

\section{Experiments}
\label{section:experiments}

\subsection{Data Collection and Preparation}

We use the Covid-19 Radiography Database dataset \cite{chowdhury2020can,rahman2021exploring}\footnote{\url{https://www.kaggle.com/datasets/tawsifurrahman/covid19-radiography-database}} which contains chest x-ray images of four categories: covid, normal, lung opacity, and viral pneumonia. We downsample the dataset to circumnavigate class imbalance. For model training we used 800 x-ray images per category totalling 3200 images. For validation and testing, we collected 1200 and 800 total images. We resize all images to 224 $\times$ 224 pixels.  The dataset is also accompanied with feature annotation masks that show the relevant regions for each of the x-ray images collected from radiologists  \cite{rahman2021exploring}.

Even though the exact number of effected images is unknown, the dataset contains confounding regions, such as marks, texts, and timestamps in many of the images.

\subsection{Model Training}

We followed a transfer learning approach using a pre-trained MobileNetV2 model \cite{sandler2018mobilenetv2}. We chose to use MobileNetV2 because it achieved better performance at the chest x-ray images classification task at a reduced computational cost after comparison against pre-trained models available at the Keras website\footnote{\url{https://keras.io/api/applications/}}. In order for the training process to affect the GradCAM explanation outputs, we only freeze and reuse the first 50 layers of MobileNetV2 and retrain the rest of the convolutional layers with a classifier layer (256 nodes with a ReLu activation with a 50\% dropout followed by a Softmax layer with 4 nodes) that we added.

We first trained the MobileNetV2 to categorise the training set into the four classes using categorical cross entropy. It was trained for 60 epochs\footnote{The model was trained with an early stop monitoring the validation loss at a patience of five epochs and a decaying learning rate = 1e-04.} using Adam optimiser. We refer to this model as the Unrefined model. We use the Unrefined model to extract good and bad GradCAM explanations. Next, we employ our eXBL algorithm using the good and bad explanations to teach the Unrefined model to focus on relevant image regions by tuning its explanations to look like the good explanations and differ from the bad explanations as much as possible. We refer to this model as the eXBL model and it was trained for 100 epochs using the same early stopping, learning rate, and optimiser as the Unrefined model.

\section{Results}
\label{section:results}

Tables \ref{table:summary_classification} and \ref{table:category_classification} show classification performance of the Unrefined and eXBL refined models. While the average AR score of GradCAM explanations produced using the eXBL model is 0.705, the explanations of the Unrefined model score an average AR of 0.685. Sample test images, masks, GradCAM outputs, and overlaid GradCAM visualisations of both the Unrefined and eXBL models are displayed in Figure \ref{figure:sample_gradcam_results}. From the sample outputs, we observe that the eXBL model was able to produce more accurate explanations that capture the relevant image regions presented with annotation masks. However, the superior explanations of the eXBL model come with a classification performance loss on half of the categories as is summarised in Table \ref{table:category_classification}. 

\begin{table}[h]

\vskip 0.15in
\begin{center}
\begin{small}
\begin{sc}
\begin{tabular}{lcccr}
\hline
Metric & Unrefined model & eXBL \\
\hline
Accuracy    & 0.950 & 0.910\\
Precision & 0.947 & 0.912\\
Recall    & 0.945 & 0.902  \\     

\hline
\end{tabular}
\end{sc}
\end{small}
\end{center}
\vspace{-20pt}
\caption{Summary of classification performances of the Unrefined and eXBL models.}
\label{table:summary_classification}
\vspace{-10pt}
\end{table}

\begin{table}[h]
\vskip 0.15in
\begin{center}
\begin{small}
\begin{sc}
\begin{tabular}{lcccr}
\hline
Category & Unrefined model & eXBL \\
\hline
Covid    & 0.925& 0.855\\
Normal & 0.930 & 0.955\\
Lung opacity    & 0.955 & 0.955  \\
Viral pneumonia    & 0.975 & 0.945       \\

    

\hline
\end{tabular}
\end{sc}
\end{small}
\end{center}
\vspace{-20pt}
\caption{Classification performance into four categories.}
\label{table:category_classification}
\vspace{-10pt}
\end{table}


\section{Conclusion}
\label{section:discussion_and_conclusion}

In this work, we have presented an approach to simplify the demanding task of feature annotation in XBL to an identification of only two model explanations. Our approach, Exemplary eXplanation-based Learning (eXBL) can tune a model's attention to focus on relevant image regions, thereby improving the saliency-based model explanations. We believe our approach is domain transferable and shows potential for real-world implementation of interactive learning using XBL.

Even though the eXBL model achieved comparable classification performance when compared against the Unrefined model (especially in categorising the Normal and Lung opacity categories, in which it scored better and equal to the Unrefined model, respectively), as is presented in Tables \ref{table:summary_classification} and \ref{table:category_classification}, we observed that there is a classification performance loss when retraining the Unrefined model with eXBL to produce good explanations. We attribute this to the accuracy-interpretability trade-off.  Although the existence of this trade-off is debated \cite{rudin2019stop,dziugaite2020enforcing}, performance loss after retraining a model could mean that the initial model was exploiting confounding regions in the training instances. It could also mean that our selection of good and bad explanations may not have been optimal and that the two exemplary explanations may be degrading model performance. 

The two exemplary explanations are selected using an objective evaluation metric, AR, and an existing dataset of annotation masks. For system development and experiment purposes, we use the masks as base knowledge. Although we believe our work presents a simple approach to implement XBL on other domains, future work should involve domain experts when picking the good and bad explanations. However, when involving end users, since the pool of explanations to choose the exemplary explanations from could be large, active learning approaches should be explored to select a subset of model explanations to prompt domain experts for feedback.

\section*{Acknowledgements}This publication has emanated from research conducted with the financial support of Science Foundation Ireland under Grant number 18/CRT/6183. For the purpose of Open Access, the author has applied a CC BY public copyright licence to any Author Accepted Manuscript version arising from this submission.








\bibliographystyle{apalike}

\bibliography{imvip_Formatting_Instructions.bib} 

\end{document}